\documentclass{article}
\pdfoutput=1 
\usepackage[utf8]{inputenc}
\usepackage{authblk}
\usepackage{setspace}
\usepackage[margin=1.25in]{geometry}
\usepackage{graphicx}
\graphicspath{ {figures/} }
\usepackage{subcaption}
\usepackage{amsmath}
\usepackage{multirow}
\usepackage{array}
\usepackage{booktabs}

\usepackage[style=ieee, 
citestyle=numeric-comp,
sorting=none]{biblatex}
\title{CRT-Net: A Generalized and Scalable Framework for the Computer-Aided Diagnosis of Electrocardiogram Signals}

\author[1]{Jingyi Liu}
\author[1*]{Zhongyu Li}
\author[2]{Xiayue Fan}
\author[1]{Jintao Yan}
\author[2]{Bolin Li}
\author[1]{Xuemeng Hu}
\author[3]{Qing Xia}
\author[2*]{Yue Wu}

\affil[1]{School of Software Engineering, Xi'an Jiaotong University, Xi'an, Shaanxi, China}
\affil[2]{Department of Cardiology, The First Afﬁliated Hospital of Xi’an Jiaotong University, Xi’an, Shaanxi, China}
\affil[3]{SenseTime Research}
\affil[*]{Corresponding authors. Email: zhongyuli@xjtu.edu.cn, yue.wu@xjtu.edu.cn}

\date{}

\begin{document} 

\maketitle

\begin{abstract}
Electrocardiogram (ECG) signals play critical roles in the clinical screening and diagnosis of many types of cardiovascular diseases. Despite deep neural networks that have been greatly facilitated computer-aided diagnosis (CAD) in many clinical tasks, the variability and complexity of ECG in the clinic still pose significant challenges in both diagnostic performance and clinical applications. In this paper, we develop a robust and scalable framework for the clinical recognition of ECG. Considering the fact that hospitals generally record ECG signals in the form of graphic waves of 2-D images, we first extract the graphic waves of 12-lead images into numerical 1-D ECG signals by a proposed bi-directional connectivity method. Subsequently, a novel deep neural network, namely CRT-Net, is designed for the fine-grained and comprehensive representation and recognition of 1-D ECG signals. The CRT-Net can well explore waveform features, morphological characteristics and time domain features of ECG by embedding convolution neural network(CNN), recurrent neural network(RNN), and transformer module in a scalable deep model, which is especially suitable in clinical scenarios with different lengths of ECG signals captured from different devices. The proposed framework is first evaluated on two widely investigated public repositories, demonstrating the superior performance of ECG recognition in comparison with state-of-the-art. Moreover, we validate the effectiveness of our proposed bi-directional connectivity and CRT-Net on clinical ECG images collected from the local hospital, including 258 patients with chronic kidney disease (CKD), 351 patients with Type-2 Diabetes (T2DM), and around 300 patients in the control group. In the experiments, our methods can achieve excellent performance in the recognition of these two types of disease, i.e., more than 90.1\% accuracy, precision, sensitivity, and F1 score.
\end{abstract}


\section{Introduction}
According to the WHO statistics in 2016, cardiovascular diseases are the leading causes of death accounting for more than 31.4\% worldwide \cite{2016World}. In general, cardiovascular diseases can be categorized into hundreds of types, where some of them are severe and even fatal, such as myocardial infarction, pulmonary embolism, and aortic dissection. Accordingly, to achieve effective and reasonable treatments, doctors need to carefully examine and screen patients to identify their types of diseases. Electrocardiogram (ECG) is considered as one of the most commonly used examination tools in clinical screening and diagnosis, which can quickly assist doctors to identify and recognize many types of cardiovascular diseases, including arrhythmia, hyperkalemia and chronic kidney disease, etc. However, ECG signals generally demonstrate varieties in morphologies, waveforms and time domain characteristics. Such 1-dimension signals are not easy to recognize in many diagnostic cases, especially for doctors with limited experiences. On the other side, despite the computed tomography angiography (CTA) examination is recognized as the golden standard for the diagnosis of many types of heart diseases, it is more time-consuming and high-expensive in comparison with ECG examination, which is also not feasible in many first-aid scenarios. Therefore, accurate recognition of ECG signals plays critical roles in clinical diagnosis of cardiovascular diseases.

In recent years, computer-aided diagnosis (CAD) has been widely investigated for the analytics of medical data with different modalities. For the ECG recognition, there have been multiple CAD methods developed through advanced techniques of machine learning, signal processing, and deep learning, which focused on the tasks of drift elimination \cite{2008Model}, waveforms detection \cite{2007A,2002Detection,2013ECG}, feature extraction \cite{2014Feature,2010Feature}, and ECG classification \cite{2011Heartbeat,2014Heartbeat,2012Heartbeat}.

The CAD methods for ECG recognition can be roughly divided into two categories, i.e., hand-crafted and learning-based. The hand-crafted methods develop representational features of ECG signals through pre-defined measurements, where the measurements can be computed based on the PQRST waves~\cite{2007A}, including the RR interval standard deviation, Maximum RR interval and R wave density, etc. Generally, the ECG measurements are designed by cardiologists with many years of clinical practices. Despite the hand-crafted features have been widely applied in diagnostic scenarios with clinical consensus, they can only work well for the recognition of some specific diseases, which cannot tackle varieties of cardiovascular diseases~\cite{2013ECG,2014Feature}. In recent years, learning-based methods have been widely investigated. This kind of methods construct deep neural networks using well-annotated ECG datasets, to automatically learn deep features through the exploration of ECG characteristics. Kiranyaz et al.\cite{2016Real} proposed a simple convolution neural network(CNN) based on the MIT-BIH arrhythmia database~\cite{1997MIT}, which can integrate feature extraction and classification. For each patient, an individual and simple CNN will be trained by using relatively small common and patient-specific training data. Acharya et al.\cite{2017A} proposed a deep convolution neural network based on the MIT-BIH arrhythmia database~\cite{1997MIT} to quickly identify different types of arrhythmia. He et al.\cite{2019Automatic} proposed a deep neural network based on the the China Physiological Signal Challenge(CPSC) dataset~\cite{2018An}, which trains two deep neural network(DNN) models composed of a residual convolution module and a long and short-term memory(LSTM) layer, extracting features from the original ECG signals. Yao et al.\cite{2019Multi} proposed a convolution neural network on the CPSC dataset~\cite{2018An}, which is based on the attention mechanism for the classification of arrhythmia. In comparison with hand-crafted methods, the learning-based methods can automatically extract representational features for cardiovascular diseases without prior knowledge.

In addition to the above proposed methods, there are also multiple challenges and public repositories released in recent years for the ECG classification and recognition, e.g., MIT-BIH~\cite{1997MIT} and the China Physiological Signal Challenge~\cite{2018An}. MIT-BIH dataset~\cite{1997MIT} is internationally recognized as a standard ECG dataset, which is the most widely used arrhythmia dataset at present. According to the AAMI standard~\cite{recommended1987mit}, MIT-BIH arrhythmia dataset~\cite{1997MIT} can be divided into 5 categories, including Normal, Supraventricular ectopic beat (SVEB), Ventricular ectopic beat (VEB), Fusion beat and Unknown beat. Compared with MIT-BIH arrhythmia dataset~\cite{1997MIT}, CPSC dataset~\cite{2018An} contains 12-lead ECG records and 9 categories, including Normal, Atrial fibrillation (AF), First-degree atrioventricular block (I-AVB), Left bundle branch block (LBBB), Right bundle branch block (RBBB), Premature atrial contraction (PAC), Premature ventricular contraction (PVC), ST-segment depression (STD), and ST-segment elevated (STE).

Despite the above endeavors have been made for the recognition of ECG signals, there are still several limitations when applying current methods in the clinical screening and diagnosis of varieties of cardiovascular diseases. Firstly, the current methods cannot well represent the characteristics of ECG signals. The ECG signals can be represented under different aspects, e.g., waveforms features, morphological characteristics, time domain features, etc, where current methods employ hand-crafted features and either convolution and recurrent neural networks for the representation of pre-set lengths of ECG signals. The representation power still need to be improved. Secondly, current methods and challenges mainly focus on the recognition some specific types of cardiovascular diseases, where most of them focus on the recognition of cardiac arrhythmia~\cite{2016Real}--\cite{2019Multi}. Actually, there are also many other types of cardiovascular diseases which require CAD methods for clinical diagnosis. Particularly, some of them are hard to differentiate in ECG even by experienced cardiologists, e.g., chronic kidney disease, pulmonary embolism, aortic dissection, etc. Thirdly, most existed methods are designed and evaluated on public datasets with well prepared ECG signals. While in practical cases, the ECG systems in many hospitals record 12-leads ECG signals using waveforms of 2-D images, which cannot be directly employed for the training of 1-D deep neural networks. More generalized methods need to be investigated for the practical diagnosis fitting hospital requirements.

Taking the above limitations and challenges into account, this paper proposes an effective, generalized and scalable framework for the recognition of 12-leads ECG signals towards clinical diagnosis. Particularly, considering the fact that many hospitals records ECG signals in the form of graphic waves with 2-D images, we first propose a bi-directional connectivity method to transform 12-leads ECG images into 1-D signals, which can accurately discriminate leads crossed neighboring leads. Then, a scalable and robust deep neural network, i.e., CRT-Net, is designed for comprehensive representation of ECG signals, which integrate CNN, RNN and transformer into a generalized network. The proposed framework is validated on multiple ECG datasets, including most widely investigated public repositories of cardiac arrhythmia, and private datasets with typical and deadly cardiovascular diseases collected from hospitals in China. The main contributions of this paper can be summarized as follows:
\begin{itemize}
\item We design a generalized and scalable deep neural network (i.e., CRT-Net) for the representation and recognition of ECG signals. The deep model can well explore the waveform features, morphological characteristics, time domain features of ECG signals, by the combination of convolution blocks, bi-directional gate recurrent unit (GRU), and transformer encoders. 

\item Different from the traditional deep neural networks for ECG recognition, our model is extensible and compatible with ECG signals of different lengths. The designed CRT-Net consists of multiple adjustable blocks which can adaptively tackle the ECG signals with different duration times for varieties of hospital devices.

\item We also propose bi-directional connectivity method to automatically extract 2-D ECG images into numerical 1-D signals. Especially, the proposed method can successfully tackle the baseline drift and large amplitude problem of 12-leads ECG graphic waves, where some neighboring leads usually intersected with each other.

\item In addition to the most widely investigated cardiovascular diseases, we apply our proposed framework in practical diagnostic cases, e.g., chronic kidney disease (CKD) and Type-2 Diabetes (T2DM), where the ECG images are directly collected from hospitals. The proposed framework can achieve superior performance for the recognition of cardiovascular diseases.
\end{itemize}

The remaining paper is organized as follows: Section 2 provides the algorithm details of bi-directional connectivity for numerical ECG signal extraction and CRT-Net for ECG signal recognition. Followed by experimental results on public and private datasets, with corresponding discussions in Section 3. Finally, Section 4 concludes the paper and presents future works.

\section{Methodology}
In this part, we present our proposed framework for the recognition of ECG signals towards clinical diagnosis. Especially, we first introduce our solution for the extraction of numerical ECG signals from the 2-D ECG images. Then a newly designed deep neural network is proposed for the feature representation and cardiovascular diseases recognition of ECG signals.

\subsection{Extraction of ECG signals}
\begin{figure}[htbp]
\centerline{\includegraphics[width=1\textwidth]{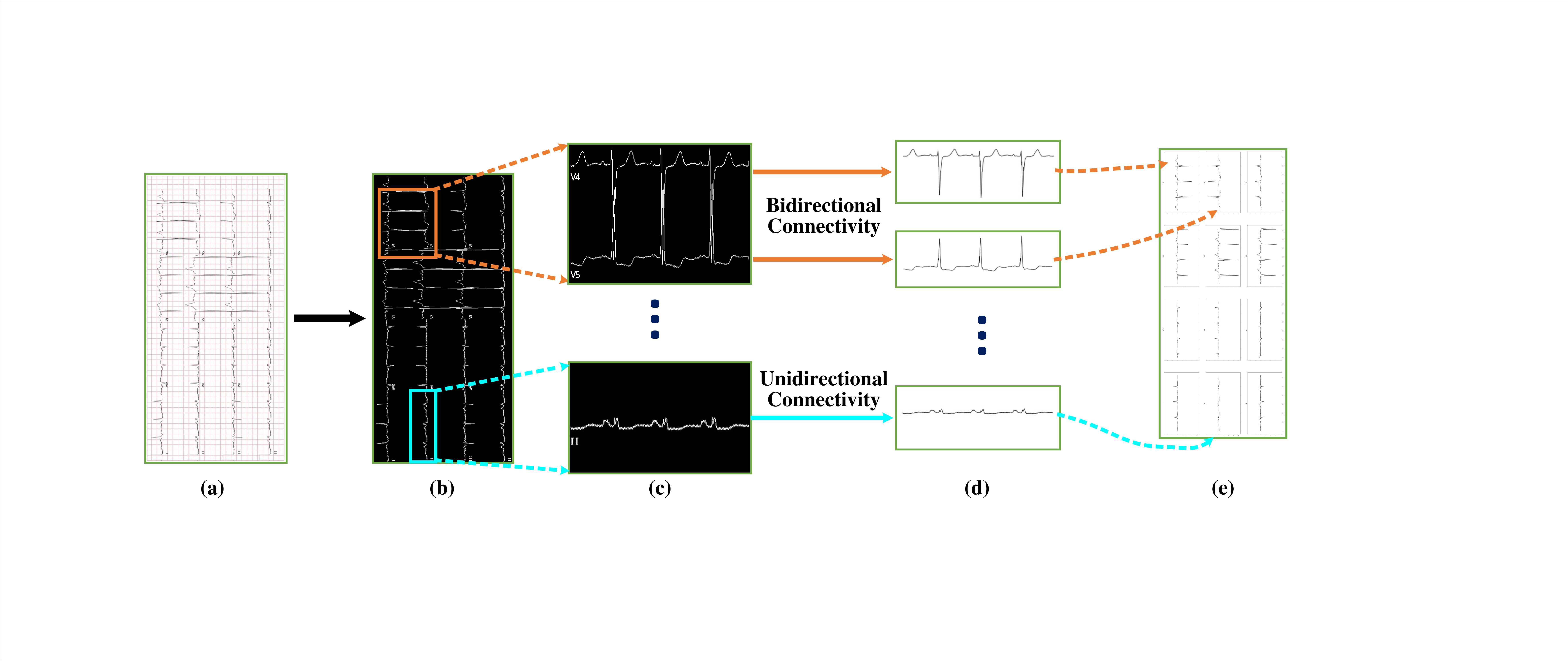}}
\caption{The workflow of ECG signal extraction from 2D images: (a) original ECG image collected from hospitals; (b) The linetype ECG waveforms after removing the background colors and grids; (c) Identify the starting and ending points of each leads, classify the leads into two categories, i.e., leads crossing with neighboring leads (top case), leads independent with neighboring leads (bottom case); (d) extraction of numerical signals for each lead, considering the crossing and independent cases respectively; (e) extraction results of 12-leads numerical ECG signals.}
\label{fig2}
\end{figure}

Considering the fact that many hospitals still record patients' ECG in the form of 2D images, it is hard directly apply deep neural networks for the recognition and differentiation of such linetype ECG images with subtle difference. Here, we first present our method for the extraction of 12-leads ECG signals in 2-D images, which can well address practical challenges in recorded data from hospitals. Fig.~\ref{fig2} presents the workflow of our method for the extraction of ECG signals from 2D images. Given an ECG image with 12-leads of waveforms (illustrated in Fig.~\ref{fig2}(a)), we need to extract the 12 black linetype waveforms with corresponding numerical times and scales, as illustrated in Fig.~\ref{fig2}(e). According to Fig.~\ref{fig2}, we divide this task into three steps: 1) removing background colors and grids (from Fig.~\ref{fig2}(a) to Fig.~\ref{fig2}(b)); 2) detecting the starting and ending points of 12-leads ECG signals respectively (from Fig.~\ref{fig2}(b) to Fig.~\ref{fig2}(c)); 3) Extraction numerical signals for each lead, considering both cross and independent leads cases respectively (from Fig.~\ref{fig2}(c) to Fig.~\ref{fig2}(d)). Here, we sequentially introduce our solution for each step.

\subsubsection{Pre-processing}
As the linetype waveforms demonstrate different colors with background grids, we can directly transform the original ECG images into gray-scale, and establish coordinate axis. Because the horizontal axis coordinates of the starting and ending points of each lead are fixed, the horizontal axis coordinates of the starting and ending points can be directly given, denoted as $x_{start}$ and $x_{end}$ respectively. The diagonal range of the ECG is given as $[x_1,y_1,x_2,y_2]$. For each ECG signal point $(x, y)$, the following conditions need to be met:
\begin{equation}
\left\{
    \begin{aligned}
        F(x,y)=1   \\
        x_1\leq x\leq x_2  \\
        y_1\leq y\leq y_2 
    \end{aligned}
    \right.
\end{equation}
where $F(x,y)$ represents the pixel value of the point $(x,y)$.

As illustrated in Fig.~\ref{fig2}(b), on the straight line $x = x_{start}$, find white pixels from top to bottom, and the first white pixel found is the starting point. For the whole lead, calculate the intersection of the line parallel to the horizontal axis and the lead image, and the line with the most intersection is the baseline. We denote the baseline as $b$.

\subsubsection{Extraction of Independent Leads}
\begin{figure}[htbp]
\centerline{\includegraphics[width=1\textwidth]{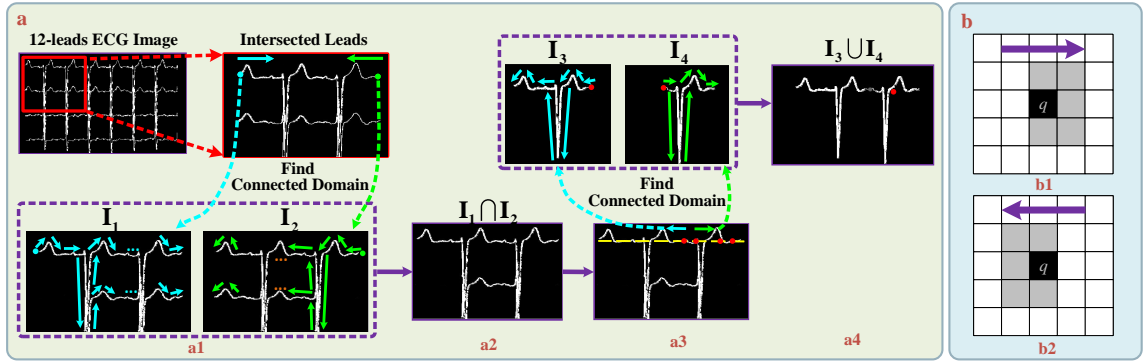}}
\caption{(a) The pipeline of our proposed bi-directional connectivity method for the numerical extraction of intersected ECG leads: (a1) given two crossed leads, the proposed method first find connected domain from both starting and ending points; (a2) the intersection of two domains ${{\bf{I}}_{\bf{1}}} \cap {{\bf{I}}_{\bf{2}}},$; (a3) Establish the baseline and searching point, then find connected domain in two neighboring searching point; (a4) the separated leads can be obtained by the iterative union of connected domains from each searching point. (b) Two directions in searching for the maximum connectivity domain: (b1) Forward directions in searching for the maximum connectivity domain; (b2)  Backward directions in searching for the maximum connectivity domain.}
\label{fig4}
\end{figure}

Because every ECG lead is continuous, we propose a method to find the maximum connectivity domain to find the ending point and extract 12-lead waves. First, we add the starting point of lead $L$ to queue $Q$, where $Q={q_1,q_2,...,q_n }, q_i=(x_i,y_i)$, and search for the maximum connectivity domain in the five forward directions on $L$, which are shown in Fig. \ref{fig4}(b1), i.e., $(x_{i+1}, y_{i-1})$, $(x_{i+1}, y_i)$, $(x_{i+1}, y_{i+1})$, $(x_i, y_{i-1})$, $(x_i, y_{i+1})$. If the pixel value of this point is 1, then add this point to the queue $Q$. Finally, traverse $Q$ until traversing all elements.

The point $ (x, y) $ is the end point $(x_{end}, y_{end})$ if it satisfies the following conditions:
\begin{equation}
   \left\{
    \begin{aligned}
        (x,y) \in Q  \\
        x=x_{end}
    \end{aligned}
    \right.
\end{equation}

The line consisting of all white pixels from the starting point to the ending point is the extracted lead $L$, which is defined as follows:
\begin{equation}
    L=\{(x,y)|F(x,y)=1,x_{start}\leq x\leq x_{end},y_{start}\leq y\leq y_{end}\}
\end{equation}

\subsubsection{Extraction of Crossed Leads}
However, baseline drift and large amplitude of some leads often appear in clinical collected ECG images (illustrated in Fig.~\ref{fig2}(c) top case). The upper and lower leads can be crossed with each other, which are hard to differentiate through the above introduced method. Here, we propose a bi-directional connectivity method to extract numerical signals from crossed leads. Firstly, we need to find the maximum connectivity domain $I_1$ in the five forward directions initialized with $(x_{start}, y_{start})$, and find the maximum connectivity domain $I_2$ in the five backward directions as shown in Fig. \ref{fig4}(b2) initialized with ($x_{end}, y_{end})$, which are shown in Fig.~\ref{fig4}(a1). Then, an logical ``OR'' operation is executed on $I_1$ and $I_2$ to get a new domain $L$ (from Fig.~\ref{fig4}(a1) to Fig.~\ref{fig4}(a2)):
\begin{equation}
    L=I_1\cap I_2
\end{equation}

In addition, we proposed the concept of searching point $S$ (illustrated in Fig.~\ref{fig4}(a3)), which is a set of the points on the baseline, defined as follows:
\begin{equation}
    S=\{(x,y)|(x,y)\in L,y=b\}
\end{equation}

Using each searching point as the starting point, we find the maximum connectivity domain $I_3$ in the five forward directions on $L$, then find the maximum connectivity domain $I_4$ in the five backward directions on $L$, which are shown in Fig.~\ref{fig4}(a3). Subsequently, we execute the logical ``AND'' operation on $I_3$ and $I_4$ to get a new domain as $L$, until the traversal is completed (from Fig.~\ref{fig4}(a3) to Fig.~\ref{fig4}(a4)):
\begin{equation}
    L=I_3\cup I_4
\end{equation}

At this time, we can get a complete lead waveform without crossover, as shown in Fig.~\ref{fig4}(a4).

Finally, for the extracted lead waves, search for white pixels from top to bottom near the baseline $b$, with the computing of distance between the longitudinal axis coordinate $y$ of the pixel and the baseline $b$. Given that the ratio of the ECG voltage to the pixel is $p$, we can calculate the converted ECG signal $N$ for each pixel, as shown below:
\begin{equation}
    N=|y-b| *p
\end{equation}

The numerical ECG signals of crossed neighboring leads can be extracted through the above bi-directional connectivity process.

\subsection{CRT-Net}
\begin{figure}[htbp]
\centerline{\includegraphics[width=1\textwidth]{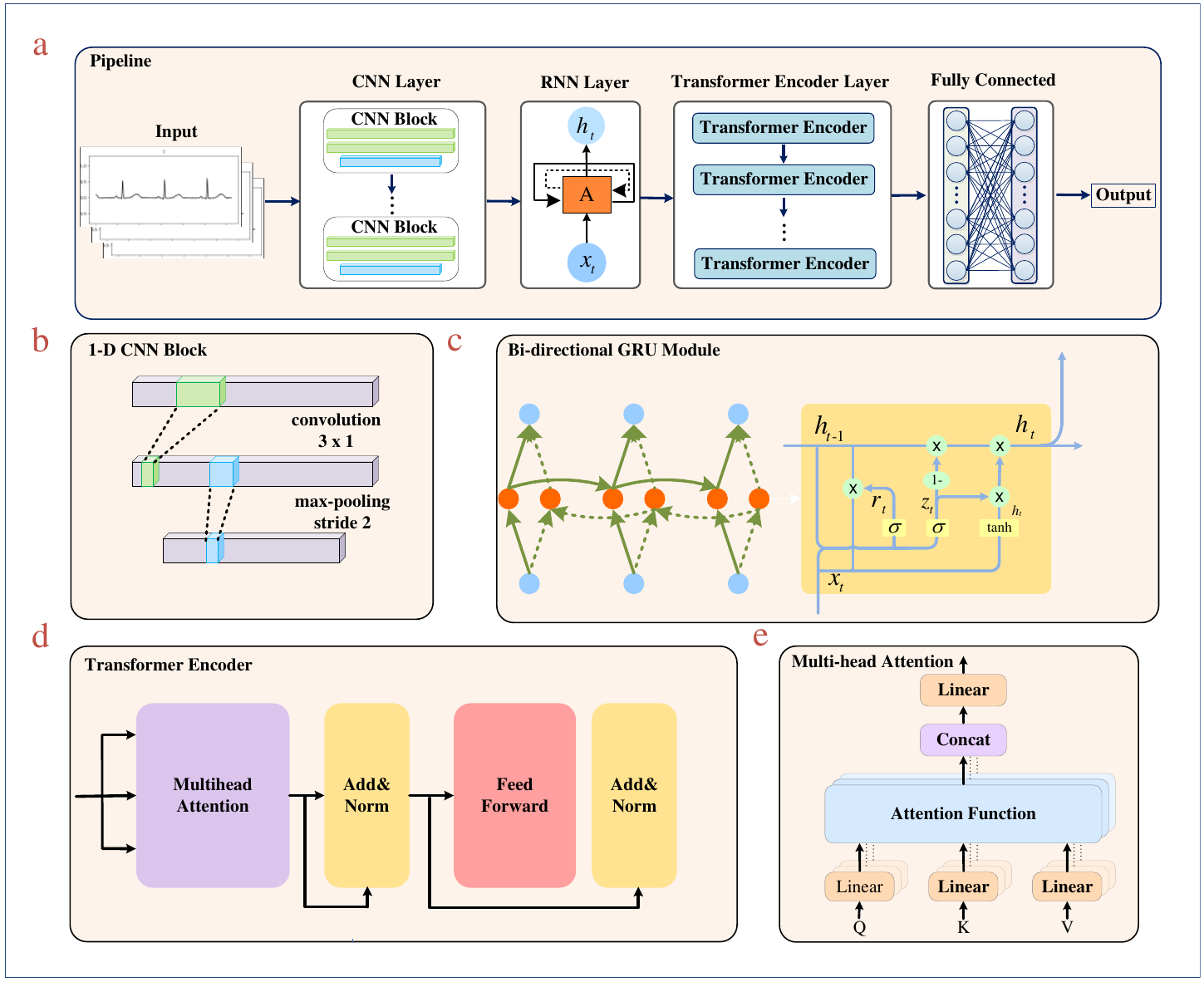}}
\caption{Architecture for CRT-Net. (a) The pipeline of CRT-Net. ECG signals are fed into CNN Blocks to extract spatial morphological features. CNN Block's output is fed into bi-directional GRU and Transformer Encoder to extract temporal features. (b) the 1-D CNN Block. (c) Internal structure and information flow during the inference of bi-directional GRU. (d) The architecture of Transformer Encoder. (e) The structure of Multi-Head Attention module, where the Multi-Head Attention module generates weights for different signal segments in different characterization subspace.}
\label{fig5}
\end{figure}

Based on the above extraction of ECG signals from 12-leads images, we propose a general and extensible deep neural network for the recognition and classification of 1-D ECG signals. Considering the fact that cardiologists generally examine ECG signals in multiple aspects, e.g., waveforms, morphologies, time domain characteristics, etc, we hope that our deep neural network can well extract the above types of features towards clinically oriented representation of ECG signals. Fig.~\ref{fig5} presents the architecture of our developed deep neural network, namely CRT-Net, which mainly includes three parts, i.e., convolution blocks, bi-directional GRU, and transformer encoders. The CRT-Net can extract ECG features and recognize different types of cardiovascular diseases end-to-end through the training of annotated ECG data.

According to Fig.~\ref{fig5}(a), given a length of 1-D ECG signals, the CRT-Net first extracts the waveform features and morphological characteristics through a series of convolution blocks. Due to the different lengths of ECG signals collected from different devices and hospitals, we encapsulate the convolution neural networks as independent blocks. When processing ECG signals with different lengths, it only needs to adjust the number of our CNN blocks. Here, we employ the VGG-Net in each CNN blocks to learn morphological features. Subsequently, since the time domain characteristics in ECG signals also need to be considered, we first introduce the recurrent neural network for the extraction of short sequential data. Especially, the bi-directional GRU module is employed for the representation of time domain features. Compared with commonly used LSTM structures, the GRU module is more efficient in the training stage, which is more likely to avoid over-fitting. This advantage is especially benefited for the clinical diagnosis of 1-D ECG signals, where the information content can be small, while the training samples are limited, in comparison with general classification tasks in natural images. Moreover, the introduced bi-directional GRU can simultaneously consider the information in ECG signals before and after a certain sequential moment. In comparison with uni-directional GRU, it can more comprehensively explore the time domain characteristics of small range of ECG signals. In addition to the exploration of morphological features and short temporal time domain characteristics, cardiologists usually need to examine ECG signals with long sequential ranges cross multiple heart beats, even through the whole lead. Here, we propose to introduce the transformer encoders for the representation of long-distance ECG signals. The transformer network architecture is first proposed in the field of natural language processing \cite{2017Attention}, which is the first 
network model to compute representations without using CNN or RNN. Therefore, to well represent the ECG signals with multiple hear beats, we further set the output of bi-directional GRU as the input of transformer encoders. To the best of our knowledge, this is the first time that integrate CNN, RNN, and transformer encoder in a extensible deep neural network for the comprehensively representation of ECG signals.

Here, we present the implementation details of each modules in the CRT-Net:

\subsubsection{Convolution Blocks}
In our CRT-Net, each convolution block includes two convolution layers and a max-pooling layer. The purpose of the convolution layer is to extract waveform features and morphological characteristics of ECG signals. In general, the first convolution layer can only extract low-level features such as edges and lines. Multi-layer convolution can extract more detailed and comprehensive features. The parameters in convolution layer are optimized by back propagation algorithm. In our model, the convolution kernel size is set as 3, where the number of convolution kernels is set as 128, with the step size of 1, and the padding of 2. Accordingly, the output and input of each convolution layer have the same size. Leaky Relu function is employed right after each convolution layer. After convolution layers, max-pooling layer is added to reduce the feature dimension and extract high-level ECG features, as shown in Fig.~\ref{fig5}(b). In our implementation, the CNN blocks can be extended with user-defined layers, according to the length of ECG signals. This setting is especially suitable for the practical collected ECG data from different devices and hospitals, where the length of each ECG signal can be range from $2.5$ seconds to $10$ seconds and even more seconds, with multiple heart beats.

\subsubsection{Bi-directional GRU}
After the extraction of waveform features and morphological characteristics in ECG signals, the CNN features can be set as input to the bi-directional GRU module, as shown in Fig.~\ref{fig5}(c). A bi-directional GRU is composed of GRUs in two directions, including both forward and backward GRUs. Given the input at current moment $t$, denoted as $x_t$, the hidden state at the previous moment and the updated gate of GRU can be denoted as $h_{t-1}$ and $z_t$ respectively, where $z_t$ determines the extent to which the state information $h_{t-1}$ at the previous time is passed to the current time $x_t$. The updated process of bi-directional GRU can be formulated as follows:

\begin{equation}
    z_t=\sigma(W_z x_t+U_z h_{t-1}+b_z)
\end{equation}
\begin{equation}
    r_t=\sigma(W_r x_r+U_r h_{t-1}+b_r)
\end{equation}
\begin{equation}
    \widetilde{h_t}=tanh(W_h x_t+U_h (h_{t-1}\otimes r_t)+b_h)
\end{equation}
\begin{equation}
    h_t=(1-z_t)\otimes h_{t-1}+z_t\otimes \widetilde{h_t}
\end{equation}
\begin{equation}
    h_t=[\overrightarrow{h_t},\overleftarrow{h_t}]
\end{equation}
where $W_z$, $W_r$, $W_h$, $U_z$, $U_r$, $U_h$ are the weight matrices, and $b_z$, $b_r$, $b_h$ are biases. $\sigma$ is the sigmoid activation function, and $\otimes$ represents the multiplication of corresponding elements. When $z_t$ is closer to 0, it means that the information of time $t$ in the hidden state of the previous layer is forgotten in the hidden layer, while when it is closer to 1, it means that the hidden layer continues to be retained. $r_t$ is the reset gate of the GRU, which controls the degree to which the state $h_{t-1}$ at the previous moment is forgotten. When $r_t$ is close to 0, it means that the information of time $t$ at the previous moment is forgotten in the current memory content $\widetilde{h_t}$. When $r_t$ is close to 1, it means that it continues to be retained in the current memory content $\widetilde{h_t}$. 

\subsubsection{Transformer Encoders}
After the bi-directional GRU, the transformer encoders are employed to extract and aggregate the time domain features of ECG signals for unified representation. The transformer encoder consists of three parts, as shown in Fig.~\ref{fig5}(d). The first is the positional encoding module. Positional encoding is used to encode the position information of the ECG signal. The formulation of the positional encoding can be listed as follows:
\begin{equation}
    PE(pos,2i)=sin(pos/10000^{2i/d_{model}})
\end{equation}
\begin{equation}
    PE(pos,2i+1)=cos(pos/10000^{2i/d_{model}})
\end{equation}
where $pos$ refers to the position of the current ECG signal point in the whole ECG signal point, and $i$ refers to the index of each finger in the ECG signal. In the even position, the sine code is used, while in the odd position, the cosine code is used. Followed by the multi-head attention module. Multi-head attention is a mechanism of executing multiple self-attention in parallel, which has the advantage of allowing the model to learn the related information in different characterization sub-spaces. The attention mechanism is to map the query to each key-value pair, where the specific calculation formula is as follows:
\begin{equation}
   Attention(Q,K,V)=softmax(\frac{QK^{T}}{\sqrt{d_k}})V
\end{equation}
where $Q$ represents the Query, $K$ represents the Key, $V$ represents the Value, and $d_k$ represents the dimension of $K$. The result of each self-attention is spliced, where the value obtained by performing a linear transformation is used as the result of multi-head attention, as shown in Fig.~\ref{fig5}(e). Finally, there is a feed forward module. We employ two fully connected layers with a corrected linear unit (Relu) function. After the multi-head attention module and the feed forward module, the layer normalization is connected, which is employed to normalize the ECG signal. Our CRT-Net includes four transformer encoders, each of which includes eight heads. After bi-directional GRU and transformer encoders, dropout is employed to randomly drop 20\% of neurons to prevent over fitting.

After the extraction of the waveform features, morphological characteristics, and time domain features of ECG signals, we use two fully connected layers to output the final classification and recognition results.

\section{Results and Discussion}
In this part, we first introduce the general experimental settings of our CRT-Net for the recognition of ECG signals, comparing CRT-Net with related state-of-the-arts on multiple public repositories. Then, we introduce the implementation details and classification results on clinical ECG images collected in hospitals.

\subsection{Experimental Settings}
In our experiment, Keras is employed for the implementation of CRT-Net. In the training phase, the maximum number of epochs is set to $100$. The early stopping mechanism is also used to alleviate over fitting. The learning rate is initially set as 0.0001, where the learning rate is iteratively reduced by 0.5 times in every 4 epochs to improve the effectiveness of the deep model. In order to reduce the influence of randomness, cross-validation is employed to compute the average as the final performance. For the ECG datasets with different lengths, only the number of user-defined CNN blocks is different.\footnote{All the codes will be available once the paper is accepted.}

In the performance evaluation, we employ four types of evaluation metrics for methods comparison, i.e., accuracy ($Acc$), precision ($Pre$), sensitivity ($Sen$), and F1 score ($F1$). The four metrics are defined as follows:
\begin{equation}
    Acc=\frac{TP+TN}{TP+TN+FP+FN}
\end{equation}
\begin{equation}
    Pre=\frac{TP}{TP+FP}
\end{equation}
\begin{equation}
    Sen=\frac{TP}{TP+FN}
\end{equation}
\begin{equation}
    F1=\frac{2*Pre*Sen}{Pre+Sen}
\end{equation}

\noindent where $TP$, $TN$, $FP$, $FN$ indicates true positive, true negative, false positive, and false negative respectively. Because of the imbalance of the sample, we calculate the weighted average as the ultimate metric to evaluate the performance of the model.

\subsection{Results on MIT-BIH Arrhythmia Dataset}
We first evaluate the proposed CRT-Net on the MIT-BIH arrhythmia dataset~\cite{1997MIT}. MIT-BIH arrhythmia dataset is the first set of commonly used standard test materials used for evaluating arrhythmia detectors, which is also the most famous and widely used arrhythmia dataset \cite{2002The}. The MIT-BIH dataset~\cite{1997MIT} includes 48 dual-channel dynamic ECG signal records collected from 47 subjects. Every ECG signal lasts more than 30 minutes, and all ECG signals are collected at a frequency of 360Hz. Every ECG signal contains two leads. In most records, the first signal is II lead obtained by placing electrodes on the chest. The second signal is usually V1, and occasionally V2, V4, or V5. According to the AAMI standard, arrhythmia heartbeats can be divided into five categories, namely N, S (SVEB), V (VEB), F, and Q, as shown in Table \ref{table1}~\cite{recommended1987mit}. We extracted a total 103,168 of ECG heartbeats based on the annotations of the MIT-BIH dataset, as shown in Table \ref{table2}. The division of training set and test set is shown in Table \ref{m}.

\begin{table}[htbp]
\centering
\caption{AAMI classes and MIT-BIH classes}
\begin{tabular}{cc}		
\hline
\specialrule{0em}{5pt}{0pt}
AAMI Classes & MIT-BIH Classes \\
\specialrule{0em}{5pt}{0pt}
\hline 
\specialrule{0em}{5pt}{0pt}
\multirow{5}*{Non ectopic beat (N)} & Normal beat \\
~ & Left bundle branch block \\
~ & Right bundle branch block \\
~ & Atrial escape beat \\
~ & Nodal escape beat \\
\specialrule{0em}{5pt}{0pt}
\multirow{4}*{Supra-ventricular ectopic beats (S)} & Atrial premature beat \\
~ & Aberrated atrial premature beat \\
~ & Nodal premature beat \\
~ & Non-conducted P-ware \\

\specialrule{0em}{5pt}{0pt}
\multirow{2}*{Ventricular epotic beats (V)} & Premature ventricular contraction \\
~ & Ventricular escape beat \\

\specialrule{0em}{5pt}{0pt}
Fusion beat (F) & Fusion of ventricular and normal beat \\

\specialrule{0em}{5pt}{0pt}
\multirow{2}*{Unclassifiable beat (Q)} & Fusion of paced and normal beat \\
~ & Unclassifiable beat \\
\specialrule{0em}{5pt}{0pt}
\hline
\end{tabular}
\label{table1}
\end{table}

\begin{table}[htbp]
\centering
\caption{The number of five classes in MIT-BIH dataset \cite{1997MIT}}
\begin{tabular}{cc}		
\hline
\specialrule{0em}{5pt}{0pt}
Classes & Number of Beats \\
\specialrule{0em}{5pt}{0pt}
\hline
\specialrule{0em}{5pt}{0pt}
N & 88521 \\
S & 2769 \\
V & 7186 \\
F & 798 \\
Q & 3894 \\
Total & 103,168 \\
\specialrule{0em}{5pt}{0pt}
\hline
\end{tabular}
\label{table2}
\end{table}

\begin{table}[htbp]
\centering
\caption{The division of training set and test set in MIT-BIH dataset \cite{1997MIT}}
\begin{tabular}{ccc}		
\toprule
Classes & Training Set & Test Set \\
\midrule
\specialrule{0em}{5pt}{0pt}
N & 79718 & 8803 \\
\specialrule{0em}{5pt}{0pt}
S & 2487 & 282 \\
\specialrule{0em}{5pt}{0pt}
V & 6422 & 764 \\
\specialrule{0em}{5pt}{0pt}
F & 725 & 73 \\
\specialrule{0em}{5pt}{0pt}
Q & 3500 & 394 \\
\specialrule{0em}{5pt}{0pt}
Total & 92,852 & 10,316 \\
\specialrule{0em}{5pt}{0pt}
\bottomrule
\end{tabular}
\label{m}
\end{table}

\begin{figure}[htbp]
\centerline{\includegraphics[width=0.5\columnwidth]{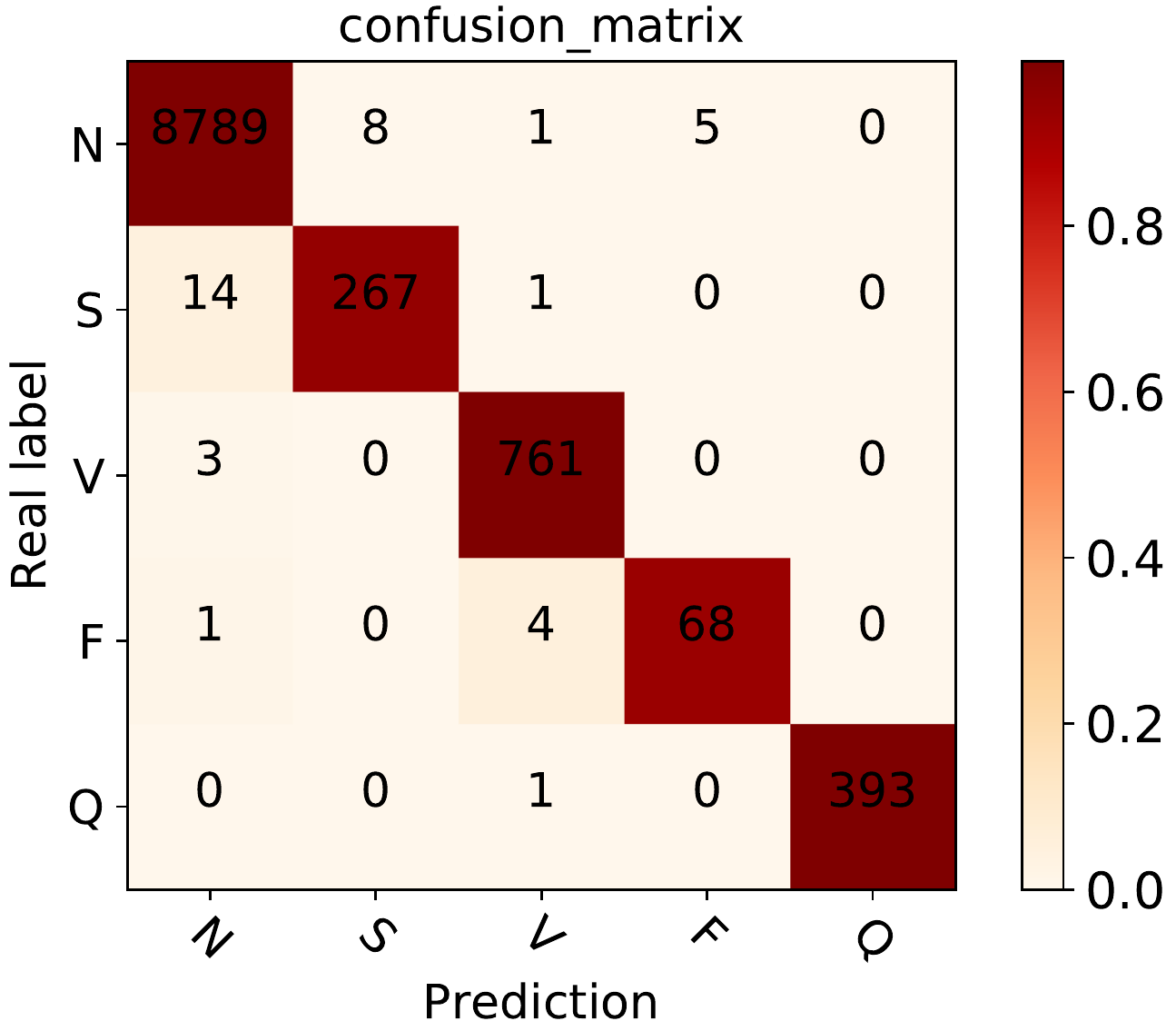}}
\caption{Confusion matrix of CRT-Net on MIT-BIH arrhythmia dataset \cite{1997MIT}}
\label{fig8}
\end{figure}

Each ECG heartbeat in the MIT-BIH dataset~\cite{1997MIT} includes about 200 ECG numerical signals. After testing different numbers of CNN Blocks, we finally chose only one CNN Block to extract the spatial morphological features of the ECG signals in this task. Fig. \ref{fig8} and Table \ref{table3} present the confusion matrix and the quantitative performance of our CRT-Net on the MIT-BIH dataset~\cite{1997MIT}, respectively. It can be seen from Table \ref{table3} that all the category metrics is above 93.2\%, which indicates that our CRT-Net can well classify and recognize different categories of arrhythmia diseases with excellent and stable performance.

\begin{table}[htbp]
\centering
\caption{the performance of CRT-Net on MIT-BIH arrhythmia dataset}
\begin{tabular}{ccccc}		
\hline
\specialrule{0em}{5pt}{0pt}
Classes & $Acc(\%)$  & $Pre(\%)$ & $Sen(\%)$ & $F1(\%)$\\
\specialrule{0em}{5pt}{0pt}
\hline
\specialrule{0em}{5pt}{0pt}
N & 99.6 & 99.8 & 99.8 & 99.8\\
S & 99.9 & 97.1 & 94.7 & 95.9 \\
V & 99.9 & 99.1 & 99.6 & 99.3 \\
F & 99.9 & 93.2 & 93.2 & 93.2 \\
Q & 100.0 & 100.0 & 99.7 & 99.9 \\
Avg & 99.6 & 99.6 & 99.6 & 99.6 \\
\specialrule{0em}{5pt}{0pt}
\hline
\end{tabular}
\label{table3}
\end{table}

In order to demonstrate the superiority of our CRT-Net on the classification of ECG signals, we compare the performance of our model with four related methods, in the task of arrhythmia heartbeat classification in five types (i.e., N, S, V, F, Q) according to the AAMI standard~\cite{recommended1987mit} introduced in Table.~\ref{table1}. The four compared methods are especially designed for the MIT-BIH dataset~\cite{1997MIT}, introduced as follows:
\begin{enumerate}
    \item 9-layer CNN\cite{2017A}: We denote the model proposed by Acharya et al. as 9-layer CNN. CNN is the most widely used model for ECG classification at present. 
    \item ABH-LSTM-CNN\cite{2019An}: Liu et al. proposed an attention-based hybrid LSTM-CNN model, namely ABH-LSTM-CNN. This model integrates CNN, LSTM and Attention in a genral framework, which can explore the overall variation trends and local features simultaneously of ECG signals.
    \item DBLSTM-WS\cite{Yildirim2018A}: Yildirim proposed a new model for deep bi-directional LSTM network-based wavelet sequences, namely DBLSTM-WS, which combines the traditional features and deep features for the representation of ECG signals.
    \item WPS\cite{Fatin2016Arrhythmia}: this model is a classical machine learning method, which is based on the wavelet transform, principal component analysis and support vector machine. We denoted this model proposed by Elhaj et al. as WPS.
\end{enumerate}

For fair comparison and avoiding randomness, we directly record the best performance of the four compared methods reported in their articles. As some of them only reported parts of evaluation metrics (e.g., only reported $Acc$), we just show the results they reported. Table.~\ref{table4} presents the performance comparison of our proposed method with four related methods for the classification of ECG signals on the MIT-BIH dataset~\cite{1997MIT}. The balanced item indicates whether this method employed strategies in tackling the unbalanced classes of ECG dataset. According to Table.~\ref{table4}, our proposed CRT-Net can achieve the best performance in comparison with other four methods, i.e., achieved 99.6\% accuracy with no additional data enhancement. These results suggests that the proposed CRT-Net is effective for the representation and classification of ECG signals from the MIT-BIH dataset~\cite{1997MIT}. Especially, in comparison with CNN-based~\cite{2017A}, RNN-based models and traditional hand-crafted features~\cite{Yildirim2018A}, our introduced transformer encoder, as well as the combination of CNN, RNN, and transformer, can more comprehensively represent and differentiate such 1-D ECG signals.

\begin{table}[htbp]
\centering
\caption{Performance comparison between CRT-Net and other methods based on MIT-BIH dataset \cite{1997MIT}}
\begin{tabular}{cccc}		
\hline
\specialrule{0em}{5pt}{0pt}
Method & Year & Balanced & Performance(\%)\\
\specialrule{0em}{5pt}{0pt}
\hline
\specialrule{0em}{5pt}{0pt}
\multirow{3}*{ 9-layer CNN\cite{2017A}} & \multirow{3}*{2017} &  \multirow{3}*{Yes} & $Acc$ : 94.0\\
 ~ & ~ & ~ & $Pre$ : 97.9 \\
 ~ & ~ & ~ & $Sen$ : 96.7 \\
\specialrule{0em}{5pt}{0pt}
\multirow{2}*{ABH-LSTM-CNN \cite{2019An}} & \multirow{2}*{2019} &\multirow{2}*{Yes} & $Acc$ : 99.3 \\
 ~ & ~ & ~ & $Sen$ : 99.6 \\
\specialrule{0em}{5pt}{0pt}
DBLSTM-WS \cite{Yildirim2018A}& 2018 & No & $Acc$ : 99.4 \\
\specialrule{0em}{5pt}{0pt}
WPS \cite{Fatin2016Arrhythmia}& 2016 & No & $Acc$ : 98.9 \\
\specialrule{0em}{5pt}{0pt}
\multirow{4}*{CRT-Net} &\multirow{4}*{2020} &  \multirow{4}*{No} & \textbf{$Acc$ : 99.6} \\
 ~ & ~ & ~ & \textbf{$Pre$ : 99.6} \\
 ~ & ~ & ~ & \textbf{$Sen$ : 99.6} \\
 ~ & ~ & ~ & \textbf{$F_1$ : 99.6} \\
\specialrule{0em}{5pt}{0pt}
\hline
\end{tabular}
\label{table4}
\end{table}

In addition, according to the suggestions of AAMI~\cite{recommended1987mit}, it is necessary to evaluate the performance of VEB (Ventricular Epotic Beats) and SVEB(Supra-ventricular Ectopic Beats) recognition. Here, we compare the CRT-Net with three related methods which also reported their performance on the recognition of VEB and SVEB. The three methods are briefly introduced as follows:
\begin{enumerate}
    \item 1-D CNN\cite{2016Real}: Kiranyaz et al. proposed an adaptive 1-D CNN for the recognition of VEB and SVEB in the MIT-BIH dataset~\cite{1997MIT}.
    \item MWT-FC\cite{Ince2009A}: We denoted the model proposed by Ince et al. as MWT-FC. This method extracts features using morphological wavelet transform, which achieve the outputs of final classification results using fully connected neural network. Wavelet transform is a traditional features extraction method in ECG classification.
    \item GRNN\cite{Li2016High}: Li et al. proposed a parallel general regression neural network to classify the heartbeat, namely GRNN. RNN is the most widely used method in temporal features extraction. 
\end{enumerate}

Table~\ref{table5} presents the performance comparison of the four methods on the recognition of VEB and SVEB on 44 records, using the consistent experimental settings. According to Table~\ref{table5}, the proposed CRT-Net can achieve excellent performance for the classification of VEB and SVEB, which also validates its effectiveness in the recognition and differentiation of different types of cardiovascular diseases.

\begin{table}[htbp]
\centering
\caption{VEB and SVEB Performance comparison between CRT-Net and other methods based on MIT-BIH dataset\cite{1997MIT}}
\begin{tabular}{ccccccc} 
\hline
\specialrule{0em}{5pt}{0pt}
{Author}  &\multicolumn{3}{c}{VEB(\%)}  & \multicolumn{3}{c}{SVEB(\%)} \\
\specialrule{0em}{5pt}{0pt}
\cmidrule(r){2-4} 
\cmidrule(r){5-7} 
\specialrule{0em}{5pt}{0pt}
   &     $Acc$ & $Sen$ & $Pre$ & $Acc$ & $Sen$ & $Pre$ \\
\specialrule{0em}{5pt}{0pt}
\hline
\specialrule{0em}{5pt}{0pt}
1-D CNN \cite{2016Real}  & 99.0 & 93.9 & 90.6 & 97.6 & 60.3 & 63.5 \\
\specialrule{0em}{5pt}{0pt}
MWT-FC \cite{Ince2009A}   & 98.3 & 84.6 & 98.7 & 97.4 & 63.5	& \textbf{99} \\
\specialrule{0em}{5pt}{0pt}
GRNN \cite{Li2016High}  &98.9 & 88 & 92.6 & 99.4 & 85.5	& 92.3 \\
\specialrule{0em}{5pt}{0pt}
CRT-Net &\textbf{99.9}& \textbf{99.6}& \textbf{99.1}&\textbf{99.9}&\textbf{94.7}&97.1 \\
\specialrule{0em}{5pt}{0pt}
\hline
\end{tabular}
\label{table5}
\end{table}

\subsection{Results on CPSC Arrhythmia Dataset}
We further evaluate the effectiveness of our proposed CRT-Net on the CPSC arrhythmia dataset~\cite{2018An} with more leads and more types of diseases. The CPSC dataset consists of 6877 12-lead ECG records, each of which has a length of 6 seconds to 60 seconds, sampled at 500Hz. These records were divided into 8 types of arrhythmia and normal sinus rhythm. Compared with the MIT-BIH arrhythmia dataset~\cite{1997MIT}, CPSC dataset~\cite{2018An} contains more leads, more diseases, and different lengths of ECG records. In addition, these records come from 11 different hospitals, which requires that the recognition methods are general and extensible for data samples from different hospitals. The detailed information of the CPSC dataset~\cite{2018An} is described in Table~\ref{table6}. The division of training set and test set is shown in Table \ref{c}.

\begin{table}[htbp]
\centering
\caption{Details of CPSC dataset, with disease types, number of records, and time durations~\cite{2018An}}
\begin{tabular}{cccccc} 
\toprule
\multirow{2}{*}{Type} & \multirow{2}{*}{Records} & \multicolumn{4}{c}{Time length(s)} \\
\cline{3-6}
\specialrule{0em}{5pt}{0pt}
   &   &  Mean & Min & Median & Max  \\
\midrule
\specialrule{0em}{5pt}{0pt}
N & 918 & 15.43 & 10.00 & 13.00 & 60.00 \\
\specialrule{0em}{5pt}{0pt}
AF & 1098 & 15.01 & 9.00 & 11.00 & 60.00 \\
\specialrule{0em}{5pt}{0pt}
I-AVB & 704 & 14.32 & 10.00 & 11.27 & 60.00 \\
\specialrule{0em}{5pt}{0pt}
LBBB & 207 & 14.92 & 9.00 & 12.00 & 60.00 \\ 
\specialrule{0em}{5pt}{0pt}
RBBB & 1695 & 14.42 & 10.00 & 11.19 & 60.00 \\ 
\specialrule{0em}{5pt}{0pt}
PAC & 556 & 19.46 & 9.00 & 4.00 & 60.00 \\ 
\specialrule{0em}{5pt}{0pt}
PVC & 672 & 20.21 & 6.00 & 15.00 & 60.00 \\ 
\specialrule{0em}{5pt}{0pt}
STD & 825 & 15.13 & 8.00 & 12.78 & 60.00 \\ 
\specialrule{0em}{5pt}{0pt}
STE & 202 & 17.15 & 10.00 & 11.89 & 60.00 \\ 
\specialrule{0em}{5pt}{0pt}
Total & 6877 & 15.79 & 6.00 & 12.00 & 60.00 \\ 
\specialrule{0em}{5pt}{0pt}
\bottomrule
\end{tabular}
\label{table6}
\end{table}

\begin{figure}[htbp]
\centerline{\includegraphics[width=0.5\columnwidth]{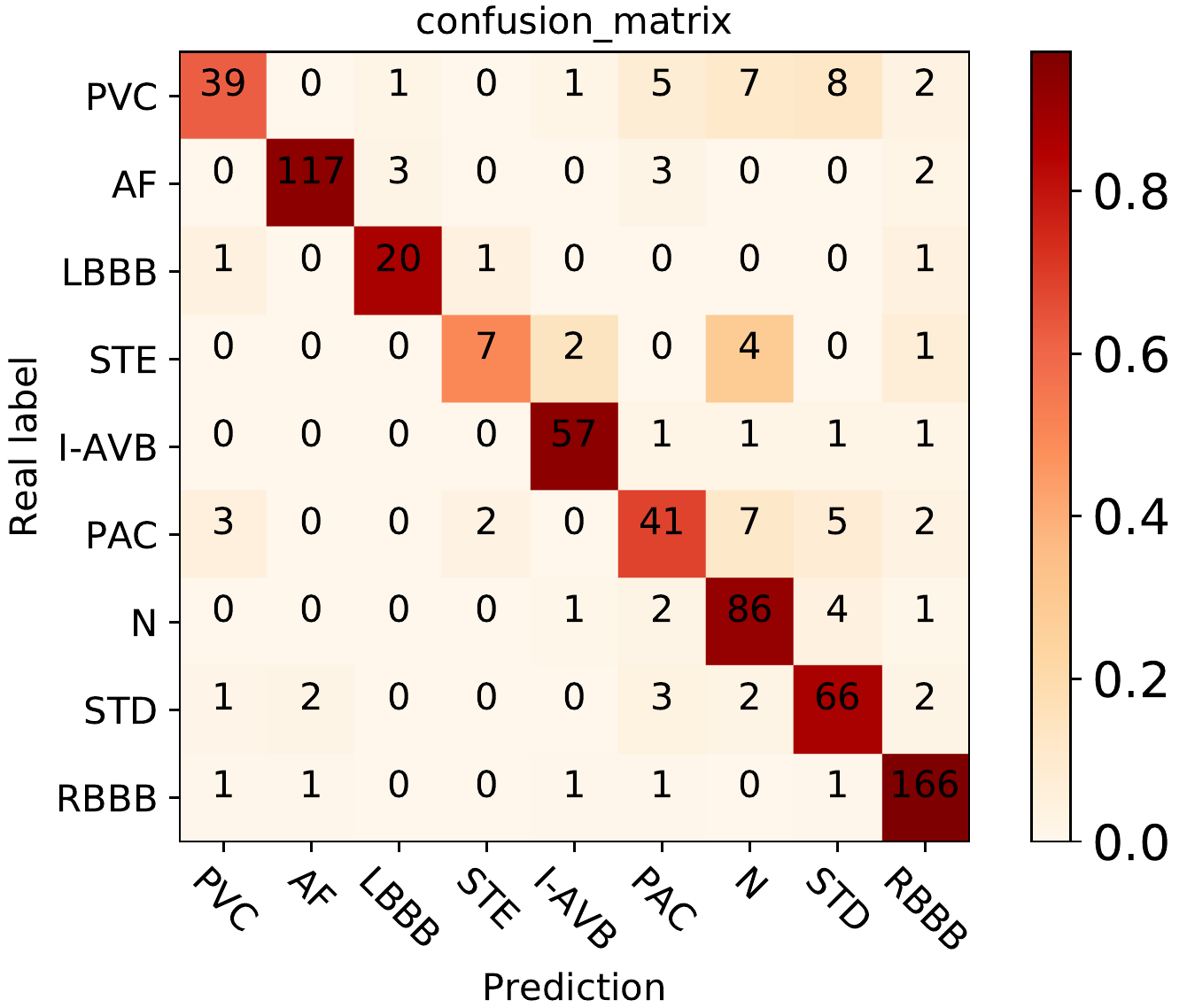}}
\caption{Confusion matrix of CRT-Net on CPSC arrhythmia dataset \cite{2018An}}
\label{fig9}
\end{figure}

\begin{table}[htbp]
\centering
\caption{The division of training set and test set in CPSC dataset \cite{2018An}}
\begin{tabular}{ccc}		
\toprule
Classes & Training Set & Test Set \\
\midrule
\specialrule{0em}{5pt}{0pt}
N & 824 & 94 \\
\specialrule{0em}{5pt}{0pt}
AF & 973 & 125 \\
\specialrule{0em}{5pt}{0pt}
I-AVB & 643 & 61 \\
\specialrule{0em}{5pt}{0pt}
LBBB & 184 & 23 \\
\specialrule{0em}{5pt}{0pt}
RBBB & 1524 & 171 \\
\specialrule{0em}{5pt}{0pt}
PAC & 496 & 60 \\
\specialrule{0em}{5pt}{0pt}
PVC & 609 & 363 \\
\specialrule{0em}{5pt}{0pt}
STD & 749 & 76 \\
\specialrule{0em}{5pt}{0pt}
STE & 188 & 14 \\
\specialrule{0em}{5pt}{0pt}
Total & 6190 & 687 \\
\specialrule{0em}{5pt}{0pt}
\bottomrule
\end{tabular}
\label{c}
\end{table}
To solve the length difference of ECG records in CPSC dataset, we adopt two strategies, i.e., cropping all ECG records into 6 seconds, extending all ECG records less than 60s into 60s. Based on the experimental results, we found that the cropping strategy not only has higher accuracy, but also has faster fitting speed. Thus, we crop all ECG records into 6 seconds. Each ECG record includes 3000 ECG numerical signals after the cropping strategy. We set 5 CNN blocks in the CRT-Net. Fig.~\ref{fig9} and Table \ref{table7} present the confusion matrix and the evaluation metrics of our CRT-Net on the CPSC dataset~\cite{2018An}, respectively. It can be seen from Table \ref{table7} that CRT-Net is effective in detecting multiple types of arrhythmia diseases under different evaluation metrics. The CRT-Net haven't achieved excellent performance in the recognition of STE, since the number of STE samples are too few.

Additionally, we also compare the CRT-Net with related methods on the CPSC dataset~\cite{2018An}. All these three compared methods reported top-ranked performance in the CPSC challenge. Here, we briefly introduce the three compared methods:
\begin{enumerate}
    \item ATI-CNN~\cite{2019Multi}: Yao et al. proposed an attention-based time-incremental convolution neural network, denoted as ATI-CNN, which can well extract ECG features by the combination of CNN, recurrent cells and attention model.
    \item Bi-LSTM~\cite{2018Classification}: Mostayed et al. proposed a recurrent neural network classifier which consists of two bi-directional LSTM layers. Bi-LSTM is suitable for exploring the characteristics of ECG signals.
    \item CRA~\cite{CHEN2020100886}: We denoted the model proposed by Chen et al. as CRA, which is based on CNN, RNN and Attention. This model won the first place in the CPSC challenge.
\end{enumerate}

\begin{table}[htbp]
\centering
\caption{the performance of CRT-Net on CPSC arrhythmia dataset \cite{2018An}}
\begin{tabular}{ccccc}		
\toprule
Classes & $Acc(\%)$  & $Pre(\%)$ & $Sen(\%)$ & $F1(\%)$\\
\midrule
\specialrule{0em}{5pt}{0pt}
N & 95.7 & 80.4 & 91.5 & 85.6\\
\specialrule{0em}{5pt}{0pt}
AF & 98.4 & 97.5 & 93.6 & 95.5  \\
\specialrule{0em}{5pt}{0pt}
I-AVB & 98.7 & 91.9 & 93.4 & 92.7\\
\specialrule{0em}{5pt}{0pt}
LBBB & 98.9 & 83.3 & 87.0 & 85.1  \\ 
\specialrule{0em}{5pt}{0pt}
RBBB & 97.5 & 93.3 & 97.1 & 95.1  \\ 
\specialrule{0em}{5pt}{0pt}
PAC & 95.4 & 73.2 & 68.3 & 70.7  \\ 
\specialrule{0em}{5pt}{0pt}
PVC & 95.6 & 86.7 & 61.9 & 72.2  \\ 
\specialrule{0em}{5pt}{0pt}
STD & 95.7 & 77.6 & 86.8 & 82.0  \\ 
\specialrule{0em}{5pt}{0pt}
STE & 98.5 & 70.0 & 50.0 & 58.3  \\ 
\specialrule{0em}{5pt}{0pt}
Avg & 87.2 & 87.3 & 87.2 & 86.9 \\
\specialrule{0em}{5pt}{0pt}
\bottomrule
\end{tabular}
\label{table7}
\end{table}

Table.~\ref{table8} presents the performance comparison between CRT-Net and other three methods on the CPSC dataset~\cite{2018An}. We record the best performance of the three compared methods reported in the challenge and their articles. Since the CPSC challenge employ F1 score as the metric for evaluation, we also record the F1 score of each method for the recognition of different types of arrhythmia diseases. According to Table.~\ref{table8}, our CRT-Net achieved the best performance in the recognition of five types of arrhythmia diseases, demonstrating superior performance in comparison with ATI-CNN~\cite{2019Multi} and Bi-LSTM~\cite{2018Classification}. These results suggests that the combination of bi-directional GRU and transformer encoder can better learn time domain features in ECG signals. Moreover, the CRA model ranked top-1 in the CPSC challenge, while the performance of our CRTNet is comparable with CRA, showing promising effectiveness in the recognition of cardiovascular diseases.

\begin{table}[htbp]
\centering
\caption{Performance comparison between CRT-Net and other methods based on CPSC dataset \cite{2018An}}
\begin{tabular}{ccccc} 
\toprule
\multirow{2}{*}{Type}  &\multicolumn{4}{c}{$F_1$(\%)} \\
\cline{2-5}
\specialrule{0em}{5pt}{0pt}
   &   ATI-CNN \cite{2019Multi} &  Bi-LSTM\cite{2018Classification} & CRA \cite{CHEN2020100886} & CRTNet \\
\midrule
\specialrule{0em}{5pt}{0pt}
N & 78.9 & 73.9 & 80.1 & \textbf{85.6}\\
\specialrule{0em}{5pt}{0pt}
AF & 92.0 & 76.8 & 93.3 & \textbf{95.5}  \\
\specialrule{0em}{5pt}{0pt}
I-AVB & 85.0 & 74.2 & 87.5 & \textbf{92.7} \\
\specialrule{0em}{5pt}{0pt}
LBBB & 87.2 & 70.6 & \textbf{88.4} & 85.1  \\ 
\specialrule{0em}{5pt}{0pt}
RBBB & 93.3 & 82.1 & 91.0 & \textbf{95.1}  \\ 
\specialrule{0em}{5pt}{0pt}
PAC & 73.6 & 59.1 & \textbf{82.6} & 70.7  \\ 
\specialrule{0em}{5pt}{0pt}
PVC & 86.1 & 80.7 & \textbf{86.9} & 72.2  \\ 
\specialrule{0em}{5pt}{0pt}
STD & 78.9 & 65.8 & 81.1 & \textbf{82.0}  \\ 
\specialrule{0em}{5pt}{0pt}
STE & 55.6 & 29.4 & \textbf{62.4} & 58.3  \\ 
\specialrule{0em}{5pt}{0pt}
\bottomrule
\end{tabular}
\label{table8}
\end{table}

\subsection{Results on Clinical ECG Images}
In this experiment, we validate the proposed ECG signal extraction and CRT-Net-based ECG recognition on clinical data collected from the Department of Cardiology, the First Affiliated Hospital of Xi’an Jiaotong University. Especially, in addition to the arrhythmia diseases, there are also many types of severe diseases which can be diagnosed through ECG images, including chronic kidney disease, type-2 diabetes, pulmonary embolism, aortic dissection, etc. Here, we evaluate the proposed framework on the recognition of chronic kidney disease (CKD) and Type-2 Diabetes (T2DM), by the collection of clinical ECG images with corresponding ground truth examination of glomerular filtration rate (GFR) and oral glucose tolerance test (OGTT) respectively. CKD is now considered as a common condition that elevates the risk of cardiovascular disease as well as kidney failure and other complications \cite{Coresh2007Prevalence}. T2DM can't be cured at present. But it can be controlled by treatment. However, if it is not treated in time, it will lead to severe complications such as diabetic ketoacidosis (DKA). Therefore, the early screening of T2DM is very important.

We collected in total $1235$ 12-leads ECG images from the First Affiliated Hospital of Xi’an Jiaotong University, which include 258 samples of CKD, 219 samples of T2DM, and around 300 samples of control group (no CKD and T2DM samples included) respectively. The ECG records in control group can be either healthy ECG records and records with other types of  cardiovascular diseases, e.g.,  myocardial infarction, pulmonary embolism, and aortic dissection. During the training phase, we extend the number of CKD, T2DM, and control group using data augmentation methods. All original samples of both CKD and T2DM are only used once in training or testing phase.\footnote{All the 12-leads ECG images will be available once the paper is accepted.}

\begin{figure}[htbp]
\centerline{\includegraphics[width=0.6\textwidth]{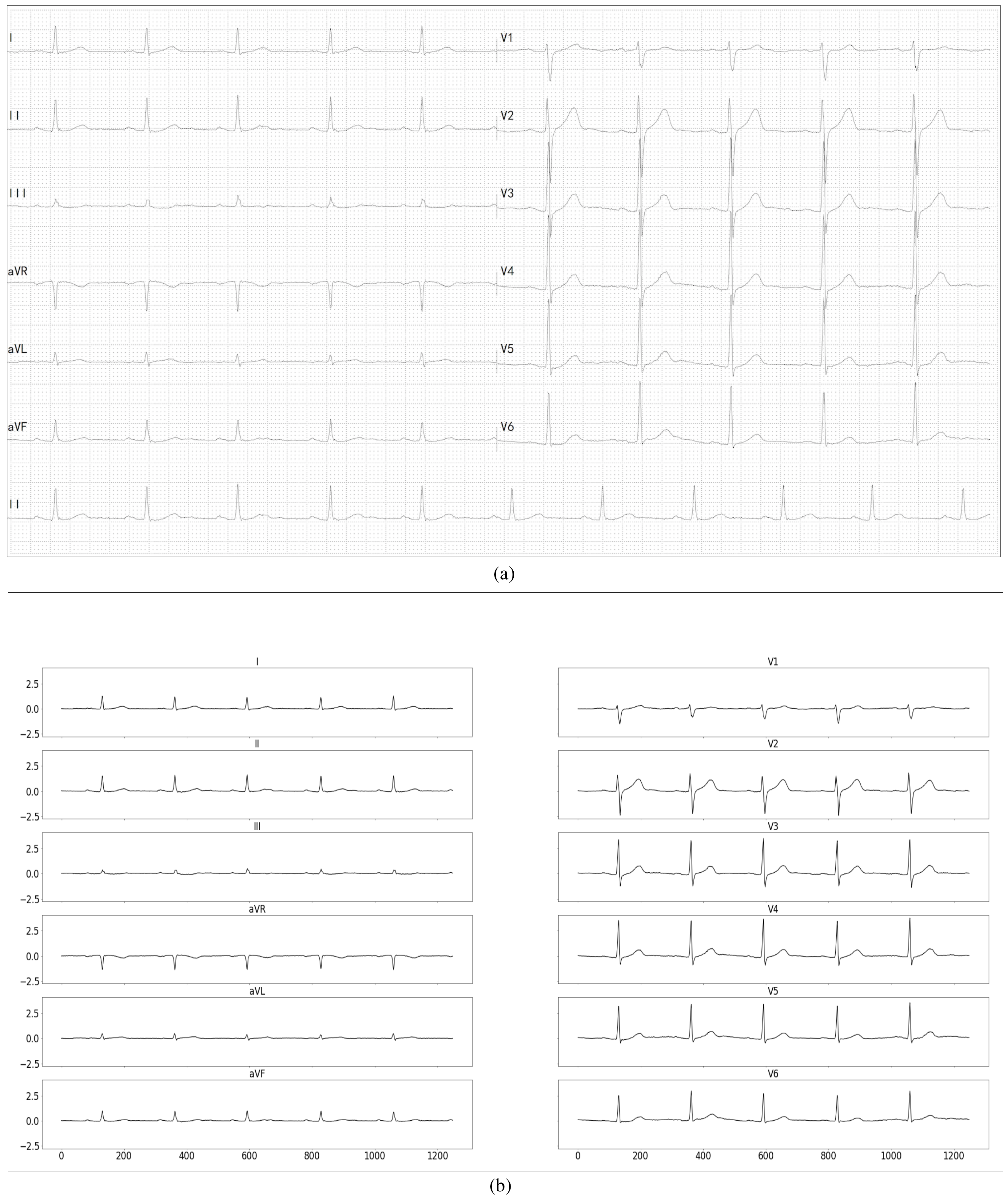}}
\caption{Visualization of extracted numerical signals based on bi-directional connectivity method: (a) 12-leads ECG images with crossed situations from the First Affiliated Hospital of Xi’an Jiaotong University; (b) extracted 12-leads ECG signals based on the bi-directional connectivity method.}
\label{fig10}
\end{figure}

We use the method based on bi-directional connectivity to extract the acquired ECG images to numerical signals. We visualized the extracted ECG signal and compare them with the original ECG images to prove the effectiveness of our method, as shown in the Fig. \ref{fig10}. Every ECG record is 5s long and sampled at 250 Hz. These ECG records are enhanced to prevent over fitting. The way of data augmentation is to intercept the ECG records randomly, filling in zeros for others. After enhanced data, there were 1548 ECG records of CKD and 1566 ECG records of non-CKD, 1446 ECG records of T2DM and 1460 ECG records of non-T2DM. These data are used to train the CRT-Net for the recognition of CKD and non-CKD, T2DM and non-T2DM respectively. The division of training set and test set is shown in Table \ref{l}.

\begin{table}[htbp]
\centering
\renewcommand\tabcolsep{3.0pt} 
\caption{The division of training set and test set on local hospital dataset}
\begin{tabular}{cccccc}		
\toprule
\multicolumn{2}{c}{Classes}  &Original Data & Augmented Data & {Training Set}  & {Test Set}\\
\midrule
\specialrule{0em}{5pt}{0pt}
\multirow{2}{*}{CKD} & N &261 & 1566 & 1253 & 313\\
\specialrule{0em}{5pt}{0pt}
 & CKD & 258 & 1548 & 1238 & 310\\
 \cmidrule(r){1-6}
\specialrule{0em}{5pt}{0pt}
\multirow{2}{*}{T2DM} & N & 365 & 1460 & 1168 & 292\\
\specialrule{0em}{5pt}{0pt}
 & T2DM &351 & 1446 & 1156 & 290\\
\specialrule{0em}{5pt}{0pt}
\bottomrule
\end{tabular}
\label{l}
\end{table}

\begin{figure}
    \centering
    \begin{subfigure}{0.4\textwidth}
        \includegraphics[width=0.9\textwidth, height=2in]{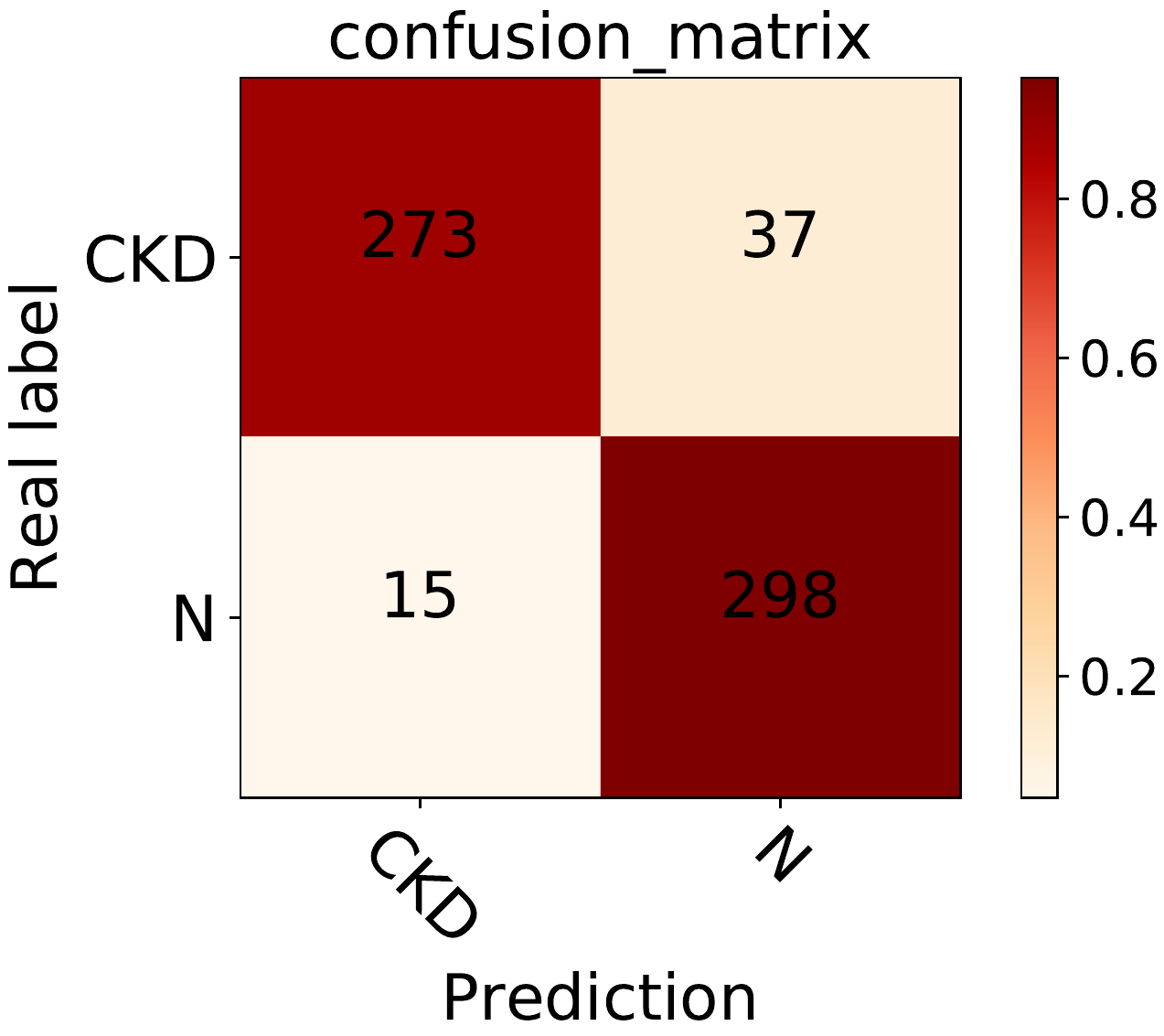}
        \caption{\label{fig:11a}}
    \end{subfigure}
    \begin{subfigure}{0.4\textwidth}
        \includegraphics[width=0.9\textwidth, height=2in]{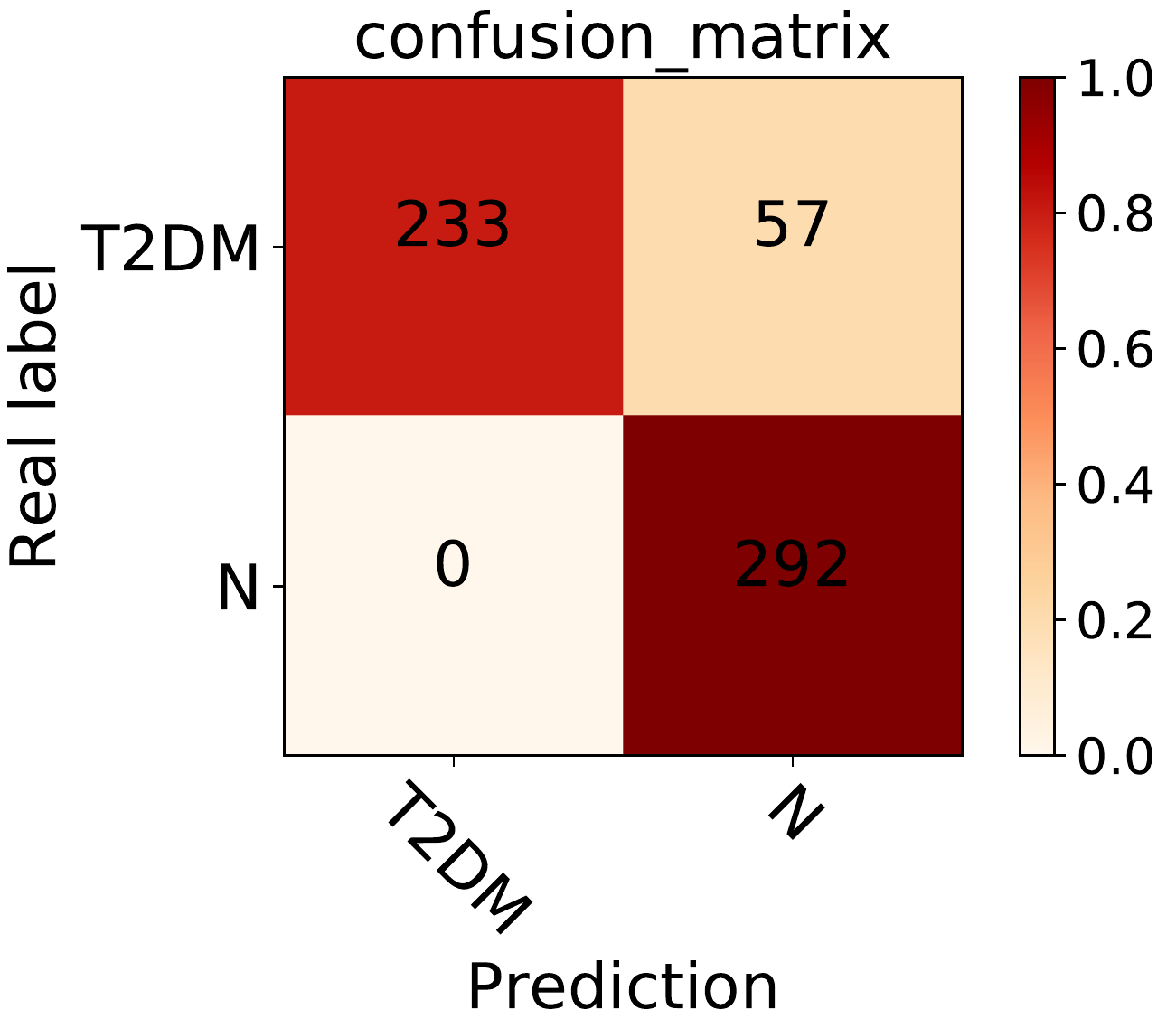}
        \caption{\label{fig:11b}}
    \end{subfigure}
    \caption{(\subref{fig:11a}) Confusion matrix of CRT-Net on CKD dataset. (\subref{fig:11b}) Confusion matrix of CRT-Net on T2DM dataset.}
    \label{fig11}
\end{figure}

Each ECG record of the hospital dataset includes 1250 ECG numerical signals. After several experiments, we finally chose 5 CNN blocks. Fig. \ref{fig11} and Table \ref{table9} show the confusion matrix and the performance of the model, respectively. These excellent results validate the effectiveness of our proposed bi-directional connectivity method for the numerical ECG extraction and the CRT-Net for ECG recognition.

\begin{table}[htbp]
\centering
\renewcommand\tabcolsep{3.0pt} 
\caption{the performance of CRT-Net on local hospital dataset}
\begin{tabular}{ccccccccc}		
\toprule
\multirow{2}{*}{Classes}  &\multicolumn{4}{c}{CKD(\%)}  &\multicolumn{4}{c}{T2DM(\%)}\\
\cmidrule(r){2-5}
\cmidrule(r){6-9}
\specialrule{0em}{5pt}{0pt}
   &  $Acc$  & $Pre$ & $Sen$ & $F1$ & $Acc$  & $Pre$ & $Sen$ & $F1$\\
\midrule
\specialrule{0em}{5pt}{0pt}
N & 91.7 & 89. & 99.1 & 95.2 & 92.0 & 83.7 & 100.0 & 91.1\\
\specialrule{0em}{5pt}{0pt}
CKD/T2DM & 91.7 & 94.8 & 88.1 & 91.3 & 90.2 & 100.0 & 80.3 & 89.1 \\
\specialrule{0em}{5pt}{0pt}
Avg & 91.7 & 91.9 & 91.7 & 91.6 & 90.2 & 91.8 & 90.2 & 90.1 \\
\specialrule{0em}{5pt}{0pt}
\bottomrule
\end{tabular}
\label{table9}
\end{table}

\begin{table}[htbp]
\centering
\caption{Performance comparison between CRT-Net and other methods based on local hospital dataset}
\renewcommand\tabcolsep{3.0pt} 
\begin{tabular}{ccccccccc} 
\toprule
\multirow{2}{*}{Approach}  &\multicolumn{4}{c}{CKD(\%)}  &\multicolumn{4}{c}{T2DM(\%)} \\
\cmidrule(r){2-5}
\cmidrule(r){6-9}
\specialrule{0em}{5pt}{0pt}
   &   $Acc$ & $Pre$ & $Sen$ & $F1$ &   $Acc$ & $Pre$ & $Sen$ & $F1$ \\
\midrule
\specialrule{0em}{5pt}{0pt}
CNN  & 84.8 & 85.2 & 84.8 & 84.1 & 83.6 & 83.8 & 83.6 & 83.6\\
\specialrule{0em}{5pt}{0pt}
Residual CNN & 80.7 & 85.9 & 80.7 & 81.5 & 80.4 & 80.2 & 80.4 & 80.3 \\
\specialrule{0em}{5pt}{0pt}
CNN-RNN & 89.4 & 90.0 & 89.2 & 89.4  & 86.2 & 86.4 & 86.2 & 86.2\\
\specialrule{0em}{5pt}{0pt}
CNN-RNN-Attention & 89.5 & 90.2 & 89.6 & 89.4 & 87.0 & 87.2 & 87.0 & 87.0 \\ 
\specialrule{0em}{5pt}{0pt}
CRTNet & \textbf{91.7} & \textbf{91.9} &\textbf{91.7}& \textbf{91.6}  & \textbf{90.2} & \textbf{91.8} & \textbf{90.2} & \textbf{90.1}\\ 
\specialrule{0em}{5pt}{0pt}
\bottomrule
\end{tabular}
\label{table10}
\end{table}

Table \ref{table10} further illustrates the reliability of our model by comparing four typical ECG classification models:
\begin{enumerate}
    \item CNN: This model refers to 9-layer CNN proposed by Acharya et al.\cite{2017A}, which consists of three convolution layers, three pooling layers and three fully connected layers.
    \item Residual CNN: This model refers to ResNet proposed by He et al.\cite{2016Deep}, which consists of five residual block. Each residual block consists of 2-layer convolution layers.
    \item CNN-RNN: This model refers to a model with a combination of convolution neural network and LSTM proposed by Verma et al.\cite{2018Cardiac}. CNN-RNN consists of five convolution layers and two LSTM layers.
    \item CNN-RNN-Attention: This model refers to CRA model proposed by Chen et al.\cite{2018Classification}. CNN-RNN-Attention consists of five convolution layers, bi-directional GRU and Attetion.
\end{enumerate}

As shown on the Table~\ref{table10}, our model can achieve superior performance in comparison with other state-of-the-arts on clinical collected data. CRT-Net not only demonstrates excellent performance on the recognition of arrhythmia diseases, but also performs good on the recognition of other diseases, such as CKD and T2DM. In the recognition of CKD and T2DM, the proposed CRT-Net is superior to the CNN and Residual CNN in terms of $ACC$, $Pre$, $Sen$ and $F1$, which shows the effectiveness of extracting time domain features. In addition, comparing to CNN-RNN and CNN-RNN-Attention, CRT-Net achieved better performance, which suggests the necessaries of transformer encodes in long ECG signals. In particular, comparing to RNN and Attention, RNN and Transformer can better extract the temporal features.

\section{CONCLUSION}
In this paper, we proposed an effective clinical-oriented framework for the recognition of ECG signals. The framework divides ECG recognition into two phases: 1) extraction of numerical ECG signals from 12-leads ECG images recorded in hospitals based on a proposed bi-directional connectivity method in tackling the crossed neighboring leads; 2) recognition and classification of ECG signals through a newly designed CRT-Net, which can comprehensively represent ECG signals by the intergration of CNN, bi-directional GRU and transformer encoders. The developed CRT-Net is compatible with input in different lengths and numbers of leads. Experimental results on 2 public datasets and 1 clinical hospital dataset validate the effectiveness of our proposed method, demonstrating superior performance in comparison with related state-of-the-arts. Accordingly, our developed framework can be implemented for the computer-aided diagnosis of ECG images in clinical cases. In the future, we focus on the exploration of interpretability of deep neural networks on the ECG data, as well as the recognition of other types of critical cardiovascular diseases.

\printbibliography

@book{2016World,
  title={World health statistics 2016: monitoring health for the SDGs, sustainable development goals},
  author={ Organization, World Health },
  pages={293-328},
  year={2016},
}

@article{2008Model,
  title={Model-Based Fiducial Points Extraction for Baseline Wandered Electrocardiograms},
  author={ Sayadi, O  and  Shamsollahi, M. B },
  journal={IEEE transactions on bio-medical engineering},
  volume={55},
  number={1},
  pages={347},
  year={2008},
}

@article{2007A,
  title={A Real-Time QRS Detection Algorithm},
  author={ Member, Senior  and IEEE and  Pan, Jiapu  and  Tompkins, Willis J. },
  journal={IEEE Transactions on Biomedical Engineering},
  volume={BME-32},
  number={3},
  pages={230-236},
  year={2007},
}

@article{2002Detection,
  title={Detection of ECG characteristic points using wavelet transforms},
  author={ Li, Cuiwei  and  Zheng, Chongxun  and  Tai, Changfeng },
  journal={IEEE Transactions on Biomedical Engineering},
  volume={42},
  number={1},
  pages={21-28},
  year={2002},
}

@article{2013ECG,
  title={ECG beat classification using PCA, LDA, ICA and Discrete Wavelet Transform},
  author={ Martis, Roshan Joy  and  Acharya, U. Rajendra  and  Min, Lim Choo },
  journal={Biomedical Signal Processing \& Control},
  volume={8},
  number={5},
  pages={437-448},
  year={2013},
}

@article{2014Feature,
  title={Feature Selection Algorithm for ECG Signals and Its Application on Heartbeat Case Determining},
  author={ Lin, L. C.  and  Yeh, Y. C.  and  Chu, T. Y. },
  journal={International Journal of Fuzzy Systems},
  volume={16},
  number={4},
  pages={483-496},
  year={2014},
}

@article{2010Feature,
  title={Feature selection algorithm for ECG signals using Range-Overlaps Method},
  author={ Yeh, Yun Chi  and  Wang, Wen June  and  Chiou, Che Wun },
  journal={Expert Systems with Applications},
  volume={37},
  number={4},
  pages={3499-3512},
  year={2010},
}

@article{2011Heartbeat,
  title={Heartbeat Classification Using Feature Selection Driven by Database Generalization Criteria},
  author={ Llamedo, M  and  Martinez, J. P },
  journal={IEEE Transactions on Biomedical Engineering},
  volume={58},
  number={3},
  pages={616-625},
  year={2011},
}

@article{2014Heartbeat,
  title={Heartbeat classification using disease-specific feature selection},
  author={ Zhang, Zhancheng  and  Dong, Jun  and  Luo, Xiaoqing  and  Choi, Kup Sze  and  Wu, Xiaojun },
  journal={Computers in Biology \& Medicine},
  volume={46},
  number={1},
  pages={79-89},
  year={2014},
}

@article{2012Heartbeat,
  title={Heartbeat Classification Using Morphological and Dynamic Features of ECG Signals},
  author={ Ye, C.  and  Vijaya Kumar, B. V. K.  and  Coimbra, M. T. },
  journal={Biomedical Engineering IEEE Transactions on},
  volume={59},
  number={10},
  pages={p.2930-2941},
  year={2012},
}

@article{2016Real,
  title={Real-Time Patient-Specific ECG Classification by 1-D Convolutional Neural Networks},
  author={ Kiranyaz, Serkan  and  Ince, Turker  and  Gabbouj, Moncef },
  journal={IEEE Transactions on Biomedical Engineering},
  volume={63},
  number={3},
  pages={664-675},
  year={2016},
}

@article{2017A,
  title={A deep convolutional neural network model to classify heartbeats},
  author={ Acharya, U Rajendra  and  Oh, Shu Lih  and  Hagiwara, Yuki  and  Tan, Jen Hong  and  Tan, Ru San },
  journal={Computers in Biology and Medicine},
  volume={89},
  year={2017},
}

@article{2019Automatic,
  title={Automatic cardiac arrhythmia classification using combination of deep residual network and bidirectional LSTM},
  author={ He, Runnan  and  Liu, Yang  and  Wang, Kuanquan  and  Zhao, Na  and  Zhang, Henggui },
  journal={IEEE Access},
  volume={PP},
  number={99},
  pages={1-1},
  year={2019},
}

@article{2019Multi,
  title={Multi-class Arrhythmia detection from 12-lead varied-length ECG using Attention-based Time-Incremental Convolutional Neural Network},
  author={ Yao, Qihang  and  Wang, Ruxin  and  Fan, Xiaomao  and  Liu, Jikui  and  Li, Ye },
  journal={Information Fusion},
  volume={53},
  year={2019},
}

@article{2017Attention,
  title={Attention Is All You Need},
  author={ Vaswani, Ashish  and  Shazeer, Noam  and  Parmar, Niki  and  Uszkoreit, Jakob  and  Jones, Llion  and  Gomez, Aidan N  and  Kaiser, Lukasz  and  Polosukhin, Illia },
  journal={arXiv},
  year={2017},
}

@article{1997MIT,
title={MIT-BIH arrhythmia database directory},
author={R.Mark and G.Moody},
journal={http://ecg.mit.edu/dbinfo.html},
year={1997},
}

@article{recommended1987mit,
title={Recommended Practice for Testing and Reporting Performance Results
of Ventricular Arrhythmia Detection Algorithms},
author={Arlington, VA, USA: Association for the Advancement of Medical Instrumentation},
year={1987},
}

@article{2002The,
  title={The impact of the MIT-BIH arrhythmia database.},
  author={ Moody, G. B.  and  Mark, R. G. },
  journal={IEEE Engineering in Medicine and Biology Magazine},
  volume={20},
  number={3},
  pages={45-50},
  year={2002},
}

@inproceedings{2019An,
  title={An Attention-based Hybrid LSTM-CNN Model for Arrhythmias Classification},
  author={ Liu, Fan  and  Zhou, Xingshe  and  Wang, Tianben  and  Cao, Jinli  and  Zhang, Yanchun },
  booktitle={2019 International Joint Conference on Neural Networks (IJCNN)},
  year={2019},
}

@article{Yildirim2018A,
  title={A novel wavelet sequences based on deep bidirectional LSTM network model for ECG signal classification},
  author={Yildirim and Ozal},
  journal={Computers in Biology \& Medicine},
  pages={189},
  year={2018},
}

@article{Fatin2016Arrhythmia,
  title={Arrhythmia recognition and classification using combined linear and nonlinear features of ECG signals},
  author={Fatin and A. and Elhaj and Naomie and Salim and Arief and R. and Harris and Tan and Tian and },
  journal={Computer Methods \& Programs in Biomedicine},
  year={2016},
}

@article{Ince2009A,
  title={A Generic and Robust System for Automated Patient-Specific Classification of ECG Signals.},
  author={Ince and Turker and Kiranyaz and Serkan and Gabbouj and Moncef},
  journal={IEEE Transactions on Biomedical Engineering},
  year={2009},
}

@article{Li2016High,
  title={High-Performance Personalized Heartbeat Classification Model for Long-Term ECG Signal},
  author={Li, Pengfei and Wang, Yu and He, Jiangchun and Wang, Lihua and Tian, Yu and Zhou, Tian Shu and Li, Tianchang and Li, Jing Song},
  journal={IEEE Transactions on Biomedical Engineering},
  pages={78-86},
  year={2016},
}

@article{2018An,
  title={An Open Access Database for Evaluating the Algorithms of Electrocardiogram Rhythm and Morphology Abnormality Detection},
  author={ Liu, Feifei  and  Liu, Chengyu  and  Zhao, Lina  and  Zhang, Xiangyu  and  Kwee, Eddie Ng Yin },
  journal={Journal of Medical Imaging and Health Informatics},
  volume={8},
  number={7},
  pages={1368-1373},
  year={2018},
}

@article{2018Classification,
  title={Classification of 12-Lead ECG Signals with Bi-directional LSTM Network},
  author={ Mostayed, Ahmed  and  Luo, Junye  and  Shu, Xingliang  and  Wee, William },
  year={2018},
}

@article{CHEN2020100886,
title = "Detection and Classification of Cardiac Arrhythmias by a Challenge-Best Deep Learning Neural Network Model",
journal = "iScience",
volume = "23",
number = "3",
pages = "100886",
year = "2020",
issn = "2589-0042",
doi = "https://doi.org/10.1016/j.isci.2020.100886",
url = "http://www.sciencedirect.com/science/article/pii/S2589004220300705",
author = "Tsai-Min Chen and Chih-Han Huang and Edward S.C. Shih and Yu-Feng Hu and Ming-Jing Hwang",
}

@article{Coresh2007Prevalence,
  title={Prevalence of Chronic Kidney Disease in the United States.},
  author={Coresh and Josef and Selvin and Elizabeth and Stevens and  Lesley, A.  and Manzi and Jane and Kusek and  John, W.  and },
  journal={JAMA: Journal of the American Medical Association},
  year={2007},
}

@article{2016Deep,
  title={Deep Residual Learning for Image Recognition},
  author={ He, Kaiming  and  Zhang, Xiangyu  and  Ren, Shaoqing  and  Sun, Jian },
  year={2016},
}

@inproceedings{2018Cardiac,
  title={Cardiac Arrhythmia Detection from Single-lead ECG using CNN and LSTM assisted by Oversampling},
  author={ Verma, Dhwaj  and  Agarwal, Sonali },
  booktitle={2018 International Conference on Advances in Computing, Communications and Informatics (ICACCI)},
  year={2018},
}
\end{document}